\pdfoutput=1 

\documentclass[10pt,twocolumn,letterpaper]{article}

\usepackage[pagenumbers]{cvpr} 
\usepackage[accsupp]{axessibility}  

\usepackage{graphicx}
\usepackage{amsmath}
\usepackage{amssymb}
\usepackage{booktabs}
\usepackage{amsmath}
\usepackage{multirow}
\usepackage{colortbl}
\usepackage{tabularray}
\usepackage{enumitem}
\usepackage{algorithm}
\usepackage{algorithmic}

\setitemize[1]{itemsep=0pt,partopsep=0pt,parsep=\parskip,topsep=0pt}

\usepackage[pagebackref,breaklinks,colorlinks]{hyperref}

\usepackage[capitalize]{cleveref}
\crefname{section}{Sec.}{Secs.}
\Crefname{section}{Section}{Sections}
\Crefname{table}{Table}{Tables}
\crefname{table}{Tab.}{Tabs.}

\def\confName{CVPR}
\def\confYear{2023}

\makeatletter
\def\thanks#1{\protected@xdef\@thanks{\@thanks
        \protect\footnotetext{#1}}}
\makeatother
\begin{document}

\title{Patch Spatio-Temporal Relation Prediction for Video Anomaly Detection}

\author{Hao Shen$^1$, Lu Shi$^1$, Wanru Xu$^1$, Yigang Cen*$^1$, Linna Zhang$^2$, Gaoyun An$^1$\\ 
$^1$Institute of Information Science, Beijing Jiaotong University, Beijing, China\\ 
$^2$School of Mechanical Engineering, Guizhou University, Guiyang, 550025, China\\
{\tt\small 22120325@bjtu.edu.cn}
}
\maketitle
\begin{abstract}
Video Anomaly Detection (VAD), aiming to identify abnormalities within a specific context and timeframe, is crucial for intelligent Video Surveillance Systems. While recent deep learning-based VAD models have shown promising results by generating high-resolution frames, they often lack competence in preserving detailed spatial and temporal coherence in video frames. To tackle this issue, we propose a self-supervised learning approach for VAD through an inter-patch relationship prediction task. Specifically, we introduce a two-branch vision transformer network designed to capture deep visual features of video frames, addressing spatial and temporal dimensions responsible for modeling appearance and motion patterns, respectively. The inter-patch relationship in each dimension is decoupled into inter-patch similarity and the order information of each patch. To mitigate memory consumption, we convert the order information prediction task into a multi-label learning problem, and the inter-patch similarity prediction task into a distance matrix regression problem. Comprehensive experiments demonstrate the effectiveness of our method, surpassing pixel-generation-based methods by a significant margin across three public benchmarks. Additionally, our approach outperforms other self-supervised learning-based methods.

\end{abstract}

\section{Introduction}
The Intelligent Video Surveillance System is designed to identify anomalous objects in real-time, encompassing both anomalous activities and anomalous entities. Anomalous objects are generally recognized as those significantly deviating from other objects within a specific environment. The demarcation between normal and abnormal objects is contingent on the context and subjects involved. Video Anomaly Detection (VAD) specializes in this task, serving as a pivotal component of an Intelligent Video Surveillance System. As the volume of video data continues to exponentially increase across diverse scenarios, VAD assumes a critical role in the realms of computer vision and pattern recognition. The advent of deep learning has led to substantial advancements in VAD. However, challenges persist due to the scarcity of anomalous activities, resulting in anomaly detection datasets containing far fewer positive samples (anomalies) than negative samples. Furthermore, some abnormal events may remain incompletely understood, even after their occurrence.

To enhance the anomaly perception of VAD models, previous studies propose to transform the anomaly detection task into a high-resolution frames generation task. These methods fall into two main categories based on the generated content: reconstruction-based and prediction-based models. As its name implies, reconstruction-based VAD methods focus on recovering input frames using generative models. In this approach, frames with high reconstruction errors are identified as anomalies during the inference stage. On the other hand, prediction-based VAD methods consider the temporal coherence between video frames. During the training stage, they aim to generate the frame at time $t$ using the preceding $t-1$ frames of a video sequence. In the inference process, anomalies are detected by examining the difference between the predicted and actual frames.

The aforementioned generation-based VAD methods exhibit promising performance. However, their effectiveness is constrained by their exclusive emphasis on the low-level pixel information of video frames, neglecting their high-level features. Additionally, the exceptional generalization ability of deep learning-based models results in well-reconstructed or predicted frames containing anomalous objects. To overcome these limitations, we introduce a novel method that leverages spatio-temporal coherence within video frames as the cornerstone for anomaly detection.

Specially, we design a inter-patch relationship prediction task as the self-supervised learning objective for VAD.

In this paper, we aim to design a more effective method that can reduce the impact of problems such as model overfitting and closed world, while also modeling the deep contextual spatio-temporal information of events to improve the accuracy of VAD.
Inspired by \cite{kim_cho_kweon_2019}, we propose a novel self-supervised learning method for VAD, by solving the random spatio-temporal patche order predicting pretext task. 
More specifically, we will divide the spatio-temporal cubes (STCs) extracted by the object extractor into patches, and then embed a random order of position encoding. The model will accomplish this by performing a self-supervised task of predicting the correct order, facilitating the modeling of events.
This challenging self-supervised task has the potential to empower the model to learn deep features of videos and capture spatio-temporal relations, thereby addressing the challenges mentioned above.

\begin{figure*}[t]
  \centering
   \includegraphics[width=0.8\linewidth]{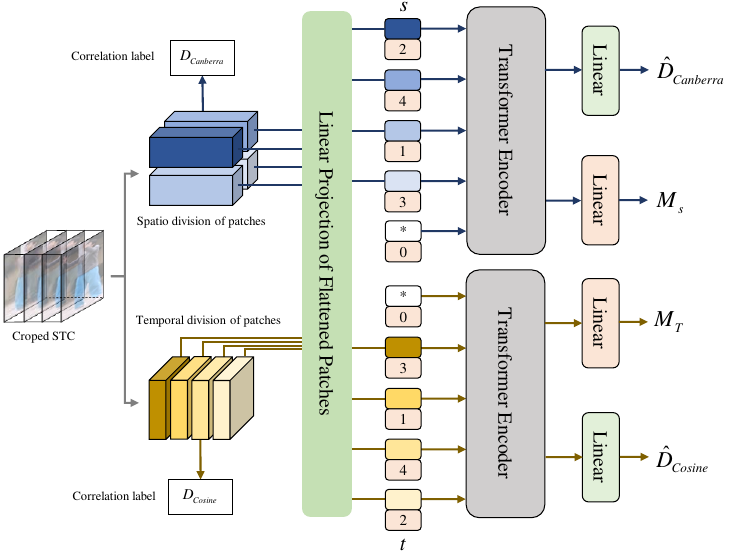}
   \caption{The framework of PSTRP. STC is divided into small patches spatially and temporally. After passing through embedding layer, randomized positional coding are embedded to these patches (The correct position order is determined by the order of colors from darkest to lightest). One vision transformer is dedicated to spatial patch order prediction and appearance feature learning (the upper transformer encoder module), while another is focused on temporal patch order prediction and motion feature capturing (the transformer encoder module in the lower part). The predictions of the model indicate the anomaly scores.}
   \label{fig:1}
\end{figure*}

To achieve it, we propose a Patche Spatio-Temporal Relation Prediction method (PSTRP) based on the two-stream Vision Transformer (ViT)\cite{dosovitskiy2020image} for VAD. we use the rearranged positions of shuffled ViT patches as labels for the self-supervised task, as illustrated in Figure \ref{fig:1}. The model outputs an order prediction matrix, as shown in Figure \ref{fig:2}.
The two-stream ViT structure of PSTRP can learn the spatial and temporal information of STC respectively through the patche order prediction task. At the same time, we use the designed distance constraint module to enable the model to predict the relations between patches at the same time, so that the model can learn spatio-temporal context features.

Our contributions are summarized as follows:
\begin{itemize}
\item PSTRP simultaneously integrates appearance and motion features to enhance the dataset for object-level anomaly detection, and address the issue of model false negatives.
\item PSTRP is the first to design a video jigsaw task tailored for ViT. It trains ViT to predict patche order, replacing reconstruction or frame prediction-based methods.
\item PSTRP introduces the distance constraint module to constrain the model in learning richer spatio-temporal information, which further enhances the novelty of anomaly detection.
\end{itemize}

\begin{figure}[t]
  \centering
   \includegraphics[width=0.8\linewidth]{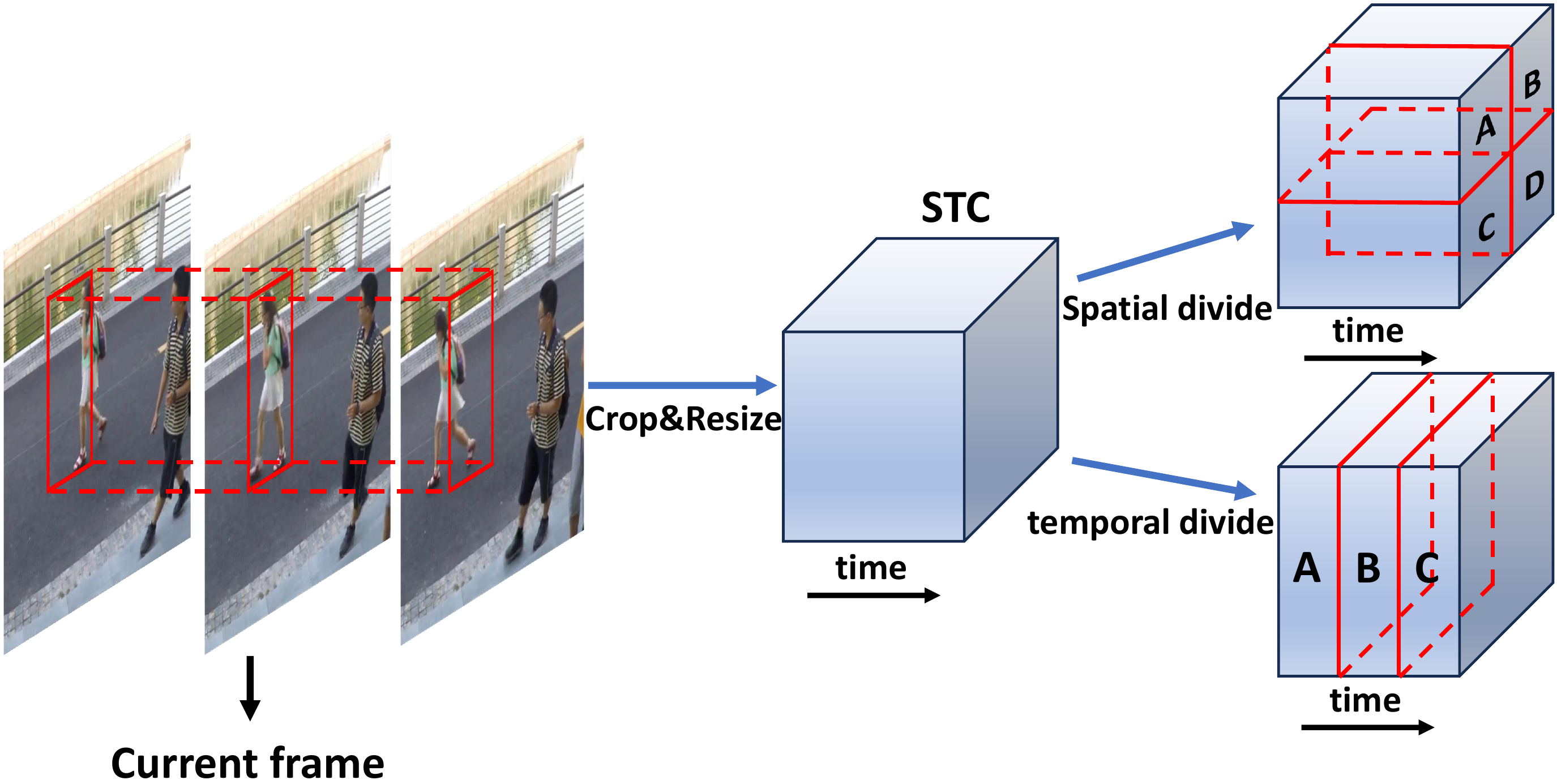}
   \caption{The extraction process of a STC and the dividing process of spatial and temporal cubes.}
   \label{fig:4}
\end{figure}
\section{Related Work}
In recent years, there have been numerous contributions in the field of deep learning for Video Anomaly Detection\cite{Cai_Zhang_Liu_Gao_Hao_2022,Zhou_Du_Zhu_Peng_Liu_Goh_2019,Wu_Liu_Shen_2019,Morais_Le_Tran_Saha_Mansour_Venkatesh_2019,Sun_Jia_Hu_Wu_2020,Nguyen_Meunier_2019,Gong_2019,chang2020clustering,park2020learning,Zhong_Chen_Jiang_Ren_2022,Park_Noh_Ham_2020,Ye_Peng_Gan_Wu_Qiao_2019,Wang_Che_Jiang_Xiao_Yang_Tang_Ye_Wang_Qi_2020,Lv_Chen_Cui_Xu_Li_Yang_2021,yang2022dynamic,tang2020integrating,liu2021hybrid,Yu_Wang_Cai_Zhu_Xu_Yin_Kloft_2020,wang2020cluster,Georgescu_2021,wang2022video}. 
These contributions can be broadly categorized into three parts: Reconstruction-Based VAD, Prediction-Based VAD and Self-Supervised Learning VAD.

\subsection{Video Anomaly Detection.}
With the development of deep learning, deep learning-based methods have achieved extraordinary results in various fields\cite{krizhevsky2012imagenet,simonyan2014very,szegedy2015going,he2016deep}. Deep learning is also used for video anomaly detection. The main difference between deep learning-based VAD and classical VAD lies in the fact that deep learning-based methods support end-to-end learning. The entire network can be trained at once, allowing for a more direct learning of task-related representations from raw data. Existing deep learning-based methods are typically categorized into the following two types.
\noindent\textbf{Reconstruction-Based Method.}
The reconstruction-based method involves training the model to learn representations of normal video behavior and using these representations for the reconstruction of video frames. Hasan \textit{et al.}\cite{Hasan_Choi_Neumann_Roy-Chowdhury_Davis_2016} utilized the extracted features as input to a fully connected neural network-based autoencoder to learn the temporal regularity in the video. Cong \textit{et al.}\cite{cong2011sparse} utilized sparse coding and dictionary learning methods to detect abnormal events by modeling normal behavior. Gong \textit{et al.}\cite{Gong_2019} used a memory-augmented autoencoder (MemAE) to improve the performance of the autoencoder based unsupervised anomaly detection methods.\\
\noindent\textbf{Prediction-Based Method.}
Methods of this kind utilize Deep Neural Networks to predict future frames, inter-frame relations, or other tasks, enabling the model to learn the spatio-temporal relations within the video. Anomalies in the video may lead to an increase in prediction errors. In pioneering works such as Liu \textit{et al.}\cite{liu2018future}, network is trained to predict future video frames. During the prediction phase, anomalies are detected by contrasting the predicted frames with the actual frames. Huang \textit{et al.}\cite{Huang_Zhao_Wang_Wu_2022} presents a framework of appearance-motion semantics representation consistency that uses the gap of appearance and motion semantic representation consistency to detect anomalies.Cao \textit{et al.}\cite{Cao_Lu_Zhang_2022} propose a novel two-stream framework, which detects the abnormal events by context recovery and knowledge retrieval.

\subsection{Self-Supervised Learning Methods.}
Self-supervised learning is a technique in which a network learns to model data by completing a pretext task with automatically generated labels. Some existing VAD works, such as \cite{Yu_Wang_Cai_Zhu_Xu_Yin_Kloft_2020}, adopt the close test as the pretext task for modeling features by training Deep Neural Networks to infer deliberately erased patches from incomplete video events. \cite{Georgescu_2021} employs multi-task self-supervised learning to enhance the learning of various feature types, including the arrow of time, motion irregularity, and knowledge distillation task.

Difference from these existing self-supervised learning based methods, we propose a novel approach for anomaly detection. Our method utilizes a patch relation prediction pretext task tailored for Vision Transformer. Unlike local perception in traditional DNNs, our method leverages the global perception capabilities of ViT, allowing for better understanding of the overall structure when dealing with the entire STC. The model learns to model appearance and motion information of the video by capturing features from patches cut in different ways. Additionally, we introduce a distance constraint module such that our method can learn deep video features and temporal relations.

\section{Method}

In the realm of unsupervised Video Anomaly Detection, the majority of approaches are focused on crafting models that characterize normal behavior, utilizing deviations as criteria for identifying anomalies. In our pursuit of a more advanced video modeling approach, we draw inspiration from \cite{kim_cho_kweon_2019} and propose a self-supervised task based on ViT patch order prediction. Specifically, we temporaly and spatialy divide the detected STC pass these divided patches through the embedding layer. Random positional embeddings are then added to these patches, and then train the model to distinctly capture the spatio-temporal features of events and accurately predict patch order. This pretext task proves to be more challenging, thereby enabling better modeling of both the temporal and spatial features of the video and providing greater global awareness compared to conventional Deep Neural Networks (DNNs). This enhanced understanding facilitates a more comprehensive grasp of the global spatio-temporal structure within the video.

\subsection{Overview}

The proposed PSTRP model consists of three major components: the object extraction module, order prediction module, and distance constraint module. As illustrated in Fig. \ref{fig:1}, the object extraction module is utilized to extract Regions of Interest (ROIs) from each frame of the video.
Subsequently, according to the positions of the ROIs in each frame, the patches in the same position of  $i$ frames before and after are cropped to form STCs.
These STCs are then spatially and temporally sliced to generate inputs for the order prediction module.
Following the embedding layer, these patches will be embedded with randomized positional coding. PSTRP is designed as a two-stream ViT, with one ViT dedicates to spatial patch order prediction and appearance feature learning. Another one focuses on temporal patch order prediction and  motion feature modeling. Throughout training, ViT is optimized to minimize prediction errors. Distance constraints are also incorporated to require ViT predicting correct inter-patch relations simultaneously, thereby  the model is constrained to learn accurate spatio-temporal context information. The model's predictions serve as anomaly scores.
During the inference phase, the model is tasked with predicting the order of patches based on the input STC. When the STC contains normal events, the model exhibits high prediction accuracy. However, its performance deteriorates significantly in the presence of anomalies. This results in small prediction errors for normal samples and significant errors for anomalies, which are then used as a criterion for anomaly detection.

\subsection{Event extraction}

Object-level anomaly detection offers a promising solution for addressing challenges associated with submerged anomalous objects\cite{Georgescu_Ionescu_Khan_Popescu_Shah_2023,Huang_Zhao_Wang_Wu_2022,reiss2022attribute,Yu_Wang_Cai_Zhu_Xu_Yin_Kloft_2020}. However, concerns arise about potential omissions of anomalous events by the model due to the incompleteness of object detection, especially when anomalous objects are not represented in the set of object categories provided in the training dataset.


To address this concern, we adopt the object extraction method proposed by \cite{Yang_Liu_Wu_Wu_Liu_2023}. Firstly, we employ the YOLOv3\cite{redmon2018yolov3} model pre-trained on the COCO dataset\cite{lin2014microsoft} to extract ROIs from the video frames, which represent appearance ROIs. Simultaneously, we compute the gradient difference between neighboring frames to obtain action ROIs. Concatenating the sets of these two types of ROIs by the extraction operation mentioned in \cite{Yang_Liu_Wu_Wu_Liu_2023} , the final video frames that encapsulate spatio-temporal ROIs information can be obtained.
Following this, for the $t^{th}$ frame, we crop patches of $i$ frames before and after at the same positions of its ROIs and resized to a same size to form the STCs. This process yields an STC of length $2i+1$, serving as an input to the model.

\subsection{Patch spatio-temporal relations prediction}
For self-supervised learning, we designed two types of pretext tasks based on ViT patches. The patch order prediction module is designed to capture the deep feature from video. The distance constrain module is designed to learn the correct patch spatio-temporal relation. Next, we will introduce these two parts in detail.

\begin{figure}[t]
  \centering
   \includegraphics[width=0.8\linewidth]{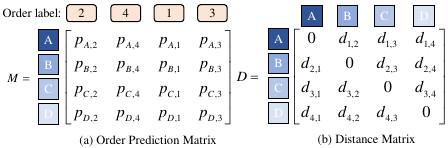}
   \caption{Relation matrix.}
   \label{fig:2}
\end{figure}

\noindent\textbf{Patch order prediction module.} 
We partition the STC along both spatial and temporal dimensions to derive spatial patches and temporal patches separately. Specifically, for a given input STC of size $L\times C\times W\times H$ (where $L$ is the length of the frame sequence, $C$ is the channel number), it can be divided into $n_s^{2}$ spatial patches of size $L\times C\times \frac{W}{n_s} \times \frac{H}{n_s}$ and $n_t$ temporal patches of size $1\times C\times W\times H$. Here,  $n_t=L$.
Subsequently, the obtained spatial patches and temporal patches are gone through the embedding layer for random positional encoding. We utilize this spatio-temporal position order as the order labels $s$ and $t$ for the patch order prediction task. The encoded temporal patches and spatial patches are separately fed into a two-stream Vision Transformer, where one stream predicts the correct spatial order, and another one predicts the temporal order of the patches. The output consists of patch order prediction matrices $M_S$ and $M_T$, which can effectively transfer a sorting task into a classification task.
In Fig. \ref{fig:2}(a), we show an $n \times n$ patch order prediction matrix,  $n\in {[2n_{s}, n_{t}]}$  (the size is determined according to the values of $n_s$ and $n_t$ in the specific task, here we show the case of $n=4$). For example,  ${p_{A,2}}$ denotes the probability that the positional encoding of the patch $A$ is $2$. Here, $A$ corresponds to the spatio-temporal cube in Fig. \ref{fig:4}, which is divided by STC in the temporal or spatial dimensions.

We employ cross-entropy loss for the order prediction task in the training process:
\begin{equation}
\begin{cases}
{\cal L}_{S}=CE(M_S,s)\\
{\cal L}_{T}=CE(M_T,t)\\
\end{cases},
\label{eq:eq1}
\end{equation}
where $M$ represents the patch order prediction matrix. $s$ and $t$ represent the labels of the patches’ order.

By completing the spatio-temporal patch order prediction task, the model can capture the deeper spatio-temporal features of videos, which will be more helpful to improve the anomaly detection accuracy.\\
\noindent\textbf{Distance constraint module.} To enhance the performance of the model, especially for the more challenging task of predicting the order of patches, we introduce a distance constraint module in this paper. In simpler terms, to handle more complex tasks effectively, we designed this module to guide the encoder in learning the accurate relations among patches.
For spatial patches, we quantify the relations distance by computing the sum of Canberra distances between edge vectors in each direction. This approach provides a representation of the distance between patches.
For temporal patches, we employ the cosine distance to depict the correlation between them. A smaller distance in this context implies a stronger correlation between patches, indicating a higher likelihood that they form a neighboring pair.
For a given STC, we establish two patch relation matrices $D_{canberra}$ and $D_{cosine}$ to  store the canberra distance and cosine distance between patches, respectively, as shown in Figure \ref{fig:2}(b). For example, $d_{1,2}$ represents the distance value between patch $1$ and patch $2$. The size is as same as the patch order prediction matrix. Obviously, $D$ is a symmetric matrix, and the elements on the diagonal are all zeroes.
The two types of relation matrices shown in Figure \ref{fig:2} jointly supervise the model to learn the correct patches' relations.
The algorithm for calculating the inter-patch relation matrices $D_{canberra}$ and $D_{cosine}$ are shown in Alg. \ref{alg:Framwork}, where
\begin{equation}
c({p_i},{p_j}) = {p_i}(h,1,c) - {p_j}( - h,1,c),
\label{eq:2}
\end{equation}
\begin{equation}
{d_{i,j}} = \sum\limits_{h = 1}^4 {\sum\limits_{c = 1}^3 {\frac{{\left| {c({p_i},{p_j})} \right|}}{{\left| {{p_i}} \right| + \left| {{p_j}} \right|}}}},
\label{eq:3}
\end{equation}
\begin{equation}
{d_{i,j}} = \frac{{{p_i} \cdot {p_j}}}{{\left| {{p_i}} \right| \cdot \left| {{p_j}} \right|}}.
\label{eq:4}
\end{equation}
\begin{algorithm}[htb]
\caption{ Framework of ensemble learning for our system.}
\label{alg:Framwork}
\begin{algorithmic}[1] 
\REQUIRE ~~\\ 
    Spatio-Temporal Object Cubes
\ENSURE ~~\\ 
    $D_{canberra}$, $D_{cosine}$
    \STATE Divide the object cubes into spatial and temporal patches of equal size, following the same procedure as the patch order prediction module;
    \STATE Create a Canberra distance matrix $D_{canberra}$ to establish pairwise correspondences between spatial patches. Additionally, establish a cosine distance matrix $D_{cosine}$ for pairwise correspondences between temporal patches;
    \STATE For each pair of spatial patches, calculate the edge vector $p_i$ in the direction  $h$ of the vector patch $i$, and subtract the edge vector $p_j$ in the opposite direction $-h$ of the spatial patch $j$. Determine the magnitude of the difference $c({p_i},{p_j})$ by the Equation \ref{eq:2} in that direction;
    \STATE Calculate the correlation distance $d_{i,j}$ using Equation \ref{eq:3} for pairs of spatial patches in each of the four directions (up, down, left, right). After computing all possible combinations of spatial patch pairs, obtain the inter-patch relation matrix $D_{canberra}$;
    \STATE Treat each pair of temporal patches as calculation vectors ${p_i}$ and ${p_j}$ and calculate the correlation distance $d_{i,j}$ using Equation \ref{eq:4} for the pair of temporal patches. After computing all possible combinations of temporal patch pairs, obtain the inter-patch relation matrix $D_{cosine}$;
\RETURN $D_{canberra}$, $D_{cosine}$. 
\end{algorithmic}
\end{algorithm}

During the training process, the matrices $D_{canberra}$ and $D_{cosine}$ serve as labels for predicting inter-patch relations. Consequently, the model is trained to capture depth relations information within the spatio-temporal context in both spatial and temporal dimensions. This enables the model to effectively model the spatio-temporal context relations inherent in the video data.
For relation prediction, we use the L2-norm loss function:
\begin{equation}
\begin{cases}
{{\cal L}_{Can}} = \left\| {{{\hat D}_{Canberra}} - {D_{Canberra}}} \right\|_2^2\\
{{\cal L}_{Cos}} = \left\| {{{\hat D}_{Cosine}} - {D_{Cosine}}} \right\|_2^2\\
\end{cases},
\label{eq:eq2}
\end{equation}
where $D$ represent the inter-patch relation matrix.
And then the total loss function can be established as follows.
\begin{equation}
{\cal L} = {\lambda _s}{{\cal L}_{S}} + {\lambda _t}{{\cal L}_{T}} + {\lambda _{can}}{{\cal L}_{Can}} + {\lambda _{\cos }}{{\cal L}_{Cos}},
\label{eq:eq3}
\end{equation}
where $\lambda$ denotes the weight of each loss function.
\subsection{Anomaly Detection on Testing Data}
When identifying anomalies, our approach aligns with \cite{wang2022video}. By ensuring the minimum values are on the diagonal, we utilize the patch order prediction matrix to derive the object-level regularity scores $r_s$ and $r_t$, as follows:
\begin{equation}
\begin{cases}
{r_s} = \min (diag({M_s}))\\
{r_t} = \min (diag({M_t}))\\
\end{cases}.
\label{eq:eq4}
\end{equation}
Hence, if the model predicts a position label incorrectly, the corresponding regularity score for that object will be small. Subsequently, we compute the frame-level regularity scores $R_s$ and $R_t$ by selecting the minimum value of the regularity scores across all objects in the frame:
\begin{equation}
\begin{cases}
{R_s} = \min ({r_{s1}},{r_{s2}},...)\\
{R_t} = \min ({r_{t1}},{r_{t2}},...)\\
\end{cases}.
\label{eq:eq5}
\end{equation}

This implies that the presence of just one anomalous object in a video frame will directly impact the regularity score for that frame. In line with \cite{wang2022video,Feng_Song_Chen_Chen_Ni_Chen_2021,Ye_Peng_Gan_Wu_Qiao_2019}, we proceed to normalize the irregularity scores for all frames within each video:

\begin{equation}
    \begin{cases}
        {R_s} = \frac{{{R_s} - \min ({R_s})}}{{\max ({R_s}) - \min ({R_s})}}\\
        {R_t} = \frac{{{R_t} - \min ({R_t})}}{{\max ({R_t}) - \min ({R_t})}}
    \end{cases}.
\end{equation}

The regularity scores obtained from the two streams (spatial patch order prediction and temporal patch order prediction) are weighted and combined through summation to yield the ultimate regularity score $R$:
\begin{equation}
    R = {\omega _s}*{R_s} + {\omega _t}*{R_t}
    \label{eq:eq6}.
\end{equation}

Here, $\omega_s$ and $\omega_t$ are set to 0.5 to ensure that both prediction tasks equally contribute to the global regularity score. Consequently, if either of the prediction tasks is not performed accurately, it has an impact on the overall regularity score.

The ultimate anomaly score $S$ is then calculated as follows:
\begin{equation}
    S = 1 - R
    \label{eq:eq7}.
\end{equation}

\section{Experiment}
\subsection{Datasets}
Our experiments were carried out on three public video anomaly detection datasets. Both training and testing sets are defined for each dataset, with anomalous events exclusively included during testing. It is noteworthy that all datasets were gathered outdoors.

\noindent\textbf{UCSD Ped2.}\cite{5539872}
The Ped2 dataset comprises 16 normal training videos and 12 test videos at a resolution of 240 × 360 pixels. The videos are captured from a fixed scene with a camera positioned above and pointed downward. Training video clips exclusively feature normal pedestrian behavior, such as walking. Abnormal events in the dataset include instances of bikers, skateboarding, and cars.

\noindent\textbf{CUHK Avenue.}\cite{lu2013abnormal}
The Avenue dataset includes 16 normal training videos and 21 test videos at a resolution of 360 × 640 pixels. The videos are collected from a fixed scene using a ground-level camera. The training video clips contain only normal behavior, while abnormal events encompass activities like throwing objects, loitering, running, movement in the wrong direction, and the presence of abnormal objects.

\noindent\textbf{ShanghaiTech Campus.}\cite{liu2018ano_pred}
The ShanghaiTech dataset stands out as the largest publicly available dataset for Video Anomaly Detection. It consists of 330 training videos and 107 test videos from 13 different scenes, all at a resolution of 480 × 856 pixels. This dataset presents challenges with complex light conditions and camera angles. Anomalies within the dataset encompass robberies, jumping, fights, car invasions, and bike riding in pedestrian areas.

\subsection{Implementation Details}
We utilize YOLOv3 pretrained on COCO for extracting object bounding boxes with the object extraction module. To exclude objects with low confidence levels, we adopt the configurations, for Ped2, Avenue, and ShanghaiTech datasets, we set confidence thresholds of 0.5, 0.8, and 0.8, respectively. These confidence thresholds remain consistent across both the training and test sets.

\noindent\textbf{Evaluation Metrics.}
Following the widely used evaluation metrics in the field of VAD, we concatenate all the frames in dataset and compute the overall frame-level Area Under the Receiver Operating Characteristic curve (AUROC) to evaluate the performance of our proposed method.

\noindent\textbf{Training Details.} 
In the training phase, we resize each object's ROI to a size of $64 \times 64$, while ensuring that the pixel values in all frames are normalized to the range $[0,1]$. Our method achieves optimal results with $L=7$ on Ped2 and Avenue, and $L=9$ on STC. For the weights of the loss function, we set $\lambda s = \lambda t = 1$ and $\lambda{can} = \lambda{cos} = 0.1$. The regularity score weights are established as $\omega_s = \omega_t = 0.5$. We employ the Adam optimizer\cite{kingma2014adam} for training, with weight decay and settings $\beta_1 = 0.9$ and $\beta_2 = 0.99$. The initial learning rate is set to $1 \times 10^{-4}$ for Ped2 and Avenue, and $2 \times 10^{-4}$ for ShanghaiTech. Training epochs are configured at 50, 100, and 100 for Ped2, Avenue, and ShanghaiTech, respectively, with a batch size of 96. The model is trained on a single NVIDIA RTX 3090 GPU.

\begin{table}
\centering
\arrayrulecolor{black}
\begin{tabular}{c|l|c|c|c} 
\hline
\multicolumn{2}{c|}{~~~~~~~~Methods}       & Ped2   & Avenue & SHTech  \\ 
\hline
\multirow{6}{*}{\rotatebox{90}{Others}}       & AnomalyNet\cite{Zhou_Du_Zhu_Peng_Liu_Goh_2019}   & 94.9 & 86.1 & N/A       \\
& SCL\cite{Lu_Shi_Jia_2013}           & N/A & 80.9    & N/A       \\
                         & Unmasking\cite{Ionescu_Smeureanu_Alexe_Popescu_2017}       & 82.2 & 80.6 & N/A       \\
                         & DeepOC\cite{Wu_Liu_Shen_2019}              & 96.9 & 86.6 & N/A       \\ 
                         & Scene-Aware\cite{Sun_Jia_Hu_Wu_2020}    & N/A & 89.6 & 74.7       \\ 
                         & MPED-RNN\cite{Morais_Le_Tran_Saha_Mansour_Venkatesh_2019}   & N/A & N/A & 73.4       \\

\hline
\multirow{8}{*}{\rotatebox{90}{Reconstruction}}      & Conv-AE\cite{Hasan_Choi_Neumann_Roy-Chowdhury_Davis_2016}   & 90.0 & 70.2 & 60.9    \\
                         & ConvLSTM-AE\cite{Luo_Liu_Gao_2017}     & 88.1 & 77.0 & N/A       \\\
                           & MemAE\cite{Gong_2019}           & 94.1 & 83.3 & 71.2    \\
                         & Stacked RNN\cite{Luo_Liu_Gao_2017_a}     & 92.2 & 81.7 & 68.0    \\
                         
                        & MNAD\cite{park2020learning}            & 90.2 & 82.8 & 69.8    \\
                         & AMC\cite{Nguyen_Meunier_2019}             & 96.2 & 86.9 & N/A       \\

                         & CDDA\cite{chang2020clustering}           & 96.5 & 86.0 & 73.3    \\

                         & Zhong et al.\cite{Zhong_Chen_Jiang_Ren_2022}           & 97.7 & 88.9 & 70.7    \\ 
                         
\hline
\multirow{7}{*}{\rotatebox{90}{Prediction}} & FFP\cite{liu2018future}              & 95.4 & 84.9 & 72.8    \\
                         & AnoPCN\cite{Ye_Peng_Gan_Wu_Qiao_2019}          & 96.8 & 86.2 & 73.6    \\
                        & AMMC-Net\cite{Cai_Zhang_Liu_Gao_Hao_2022}           & 96.9 & 86.6 & 73.7    \\
                         & MNAD\cite{Park_Noh_Ham_2020}           & 97.0 & 88.5 & 70.5    \\
                         & ROADMAP\cite{Wang_Che_Jiang_Xiao_Yang_Tang_Ye_Wang_Qi_2020}         & 96.3 & 88.3 & 76.6    \\
                         & MPN\cite{Lv_Chen_Cui_Xu_Li_Yang_2021}           & 96.9 & 89.5 & 73.8    \\

                         & DLAN-AC\cite{yang2022dynamic}        & 97.6 & 89.9 & 74.7   \\
\hline
\multirow{7}{*}{\rotatebox{90}{Hybrid}}
    &ST-CAE\cite{zhao2017spatio} &91.2 &80.9 &N/A  \\
    &MPED-RNN\cite{morais2019learning} &N/A &N/A &73.4\\
    &AnoPCN\cite{ye2019anopcn} &96.8 &86.2 &73.6\\
    &IntegradAE\cite{tang2020integrating} &96.8 &86.2 &73.6\\
    &HF$^2$-VAD\cite{liu2021hybrid} &\textbf{99.3} &91.1 &76.2\\
    &VEC-A\cite{Yu_Wang_Cai_Zhu_Xu_Yin_Kloft_2020}&96.9 &90.2 &74.7\\
    &VEC-AM\cite{Yu_Wang_Cai_Zhu_Xu_Yin_Kloft_2020}&97.3 &89.6 &74.8\\
\hline
\multirow{3}{*}{\rotatebox{90}{SSL}} & CAC\cite{wang2020cluster}              & N/A  & 87.0 & 79.3    \\
                         & SS-MTL\cite{Georgescu_2021}          & 97.5 & 91.5 & \underline{82.4}    \\
                         & Jigsaw*\cite{wang2022video}          & 98.2 & \underline{91.6} & \textbf{83.4}    \\

\hline
                         & \textbf{PSTRP}       & \underline{98.7} & \textbf{92.5} & 80.4    \\
\hline
\end{tabular}
\arrayrulecolor{black}
 \caption{AUROC(\%) performance on Ped2, Avenue, and ShanghaiTech datasets. Results marked with * are reproducible results under the same computational conditions as our method (RTX 4090).}
  \label{tab:tab1}
\end{table}

\subsection{Experimental Results}
To verify the effectiveness of our proposed PSTRP framework for video anomaly detection, we compare it against various state-of-the-art methods on three benchmark datasets. Table \ref{tab:tab1} illustrates the performance of our method on Avenue achieving state-of-the-art results compared to other methods. This demonstrates the sensitivity to abnormal events and the superior ability of our method to detect the anomaly in video. For the Ped2 and ShanghaiTech datasets, although our method did not achieve the best performance, it still outperforms the vast majority of existing methods. We also have conducted an analysis to understand these results. These datasets are characterized by a larger volume, encompassing a broader spectrum of anomaly event types, and featuring more complex and diverse moving objects. In Section \ref{sec:44}, our ablation experiments on the depth of the Vision Transformer used in our approach suggest that increasing the depth has the potential to further enhance the model's performance. In the future, we can try to continue to improve the performance of our model by increasing the size of ViT.

\begin{figure*}[t]
  \centering
   \includegraphics[width=1\linewidth]{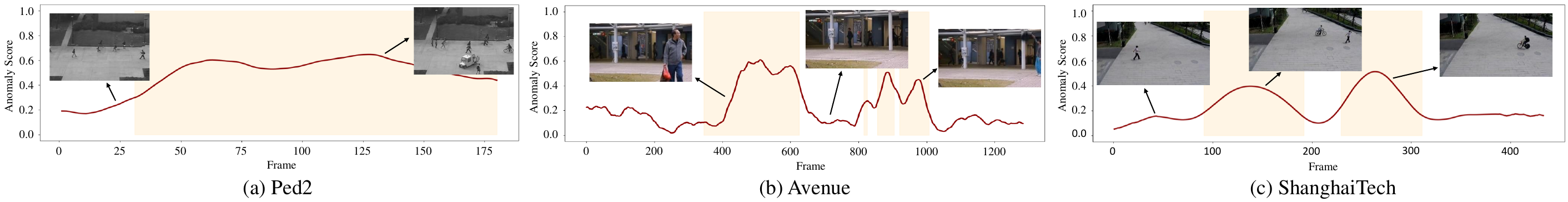}
   \caption{Illustrations of anomaly score that denotes the reconstruction error in Ped2, Avenue and SHTech datasets. Orange region in graph denotes the time sequences that abnormal situation exists in video frames. As shown in graph, anomaly scores (Red curve) dramatically increase with the high reconstruction error when the abnormal frames start.}
   \label{fig:3}
\end{figure*}

\subsection{Ablation Study and Analysis}
\label{sec:44}
\begin{table}
\centering
\begin{tabular}{ll|lll} 
\hline
OPT & DCS & Ped2          & Avenue        & SHTech   \\ 
\hline
$\times$   & $\times$   & 96.0 & 88.9 & 77.9           \\
$\times$   & $\checkmark$   & 97.4(+1.4) & 90.2(+1.3) & 80.0(+2.1)           \\
$\checkmark$   & $\times$   & 96.2(+0.2) & 91.3(+2.4) & 79.8(+1.9)           \\
$\checkmark$   & $\checkmark$   & \textbf{98.7(+2.1)} & \textbf{92.5(+3.6)} & \textbf{80.4(+2.5)}  \\
\hline
\end{tabular}
\caption{Ablation studies of each component in our PSTRP on three benchmarks. (OPT: object optimization, DCS: distance constraint)}
\label{tab:tab2}
\end{table}



\noindent\textbf{Ablation analysis for various units.}
To validate the performance enhancement of the object optimization module and distance constraint module in video anomaly detection, we conducted several sets of ablation experiments to individually assess their effectiveness.
First, we conducted ablation experiments to assess the effectiveness of the object detection optimization module and the distance constraint module. The results are presented in Table \ref{tab:tab2}. The findings in Table \ref{tab:tab2} indicate that incorporating the distance constraint module alone leads to an improvement in our method, with performance enhancements of 1.4\%, 1.3\%, and 2.1\% on the three datasets, respectively. This suggests that refining motion object detection is effective in enhancing the accuracy of video anomaly detection.
When the object detection optimization module is added alone, the model's performance improves by 0.2\%, 2.4\%, and 1.9\%, indicating that the object detection optimization module positively influences the model in correctly learning the spatio-temporal relations of the patches. This enables the model to capture deep spatio-temporal features. Combining both modules results in the best performance,  the effectiveness of our added modules is demonstrated through ablation learning. This enhancement allows the model to learn deep features and spatio-temporal relations within the video data.\\
\noindent\textbf{Various pretext tasks}
For various combinations of the number of patches, we define a range of pretext tasks and conduct experiments on three benchmark datasets. Table \ref{tab:tab3} presents the results. It can be seen that for the temporal order prediction, Ped2 and Avenue exhibit the best performance at $T=7$, while SHTech performs optimally at $T=9$. In terms of the spatial order prediction, Avenue achieves its peak performance at $S=16$, while the optimal performance is observed at $S=16$ and $S=9$  for Ped2 and SHTech, respectively.
Simultaneously, it is notable that the model's performance gradually improves with the incremental increase of $T$ and $S$ initially. However, the model encounters challenges in adaptation when the number of patches reaches a certain level due to the complexity introduced by the increased patches.\\
\begin{table}
\centering
\begin{tabular}{c|cc|ccc} 
\hline
Task& T & S & Ped2          & Avenue        & SHTech         \\ 
\hline
1        & 5        & 4      & 93.1          & 86.4          & 74.9           \\
2        & 5        & 9      & 95.7          & 90.8          & 78.2           \\
3        & 7        & 4      & 96.2          & 88.5          & 77.5           \\
4        & 7        & 9      & \textbf{98.7} & 91.6          & 79.3           \\
5        & 7        & 16     & 95.2          & \textbf{92.5} & 76.9           \\
6        & 9        & 9      & 96.7          & 89.3          & \textbf{80.4}  \\
7        & 9        & 16     & 94.3          & 90.1          & 76.0           \\
\hline
\end{tabular}
\caption{Performance in terms of AUROC(\%)  was evaluated across three datasets for varying combinations of patches quantities.}
\label{tab:tab3}
\end{table}

\noindent\textbf{Size of backbone}
We also compare the performance among backbones of different sizes. As shown in Table \ref{tab:tab4}, we employ ViT-B, ViT-L, and ViT-H as the backbone for our anomaly detection method on the three benchmark datasets, respectively. The experimental results indicate that as the size of ViT gradually increases, the AUROC also shows a gradual improvement. Due to the memory limitations of the graphics card, we did not test with a larger scale of ViT. However, we infer that the performance of our method can be further enhanced by continuing to increase the size of the backbone.
\begin{table}
\centering
\begin{tabular}{c|ccc} 
\hline
Backbone & Ped2          & Avenue & SHTech  \\ 
\hline
ViT-B/64 & 96.1          & 88.9       & 73.6        \\
ViT-L/64 & 98.3          & 90.0       & 78.5        \\
ViT-H/64 & \textbf{98.7} & \textbf{92.5}       & \textbf{80.4}        \\
\hline
\end{tabular}
\caption{AUROC performance (\%) on different sizes of ViT backbones.}
\label{tab:tab4}
\end{table}
\subsection{Qualitative Results}
In Fig. \ref{fig:3}, we visualise the anomaly scores obtained for videos containing anomalies detected by our method. We evaluate the consistency between the model's anomaly scores and the ground truth on the UCSD Ped2 and CUHK Avenue and Shanghai Tech datasets. Specifically, we evaluated the model's ability to detect cars entering and exiting the field of view in UCSD Ped2, and people with anomalous behaviours in CUHK Avenue, as well as the ability to detect cyclists riding on the pavement in Shanghai Tech. The results from Fig. \ref{fig:3} show that the anomaly scores show a sharp increase when anomalous behaviours are present, whereas the anomaly scores show low values in the time interval without anomalies, which suggests that our method is capable of identifying a wide range of anomalies and is highly robust to anomalous behaviour recognition.
Also, it shows that our method is highly sensitive to the occurrence of anomalies and is capable of accurately detecting the intervals of anomaly events.
\section{Conclusions}
In this paper, we introduced a novel self-supervised learning method based on the pretext task of predicting the patch spatio-temporal order. To achieve this, we proposed a model called PSTRP based on a two-stream vision transformer for video anomaly detection. Object detection optimization, as well as distance constraints modules, were incorporated to enable the model to capture deep video features and spatio-temporal relations, which can distinguish abnormal situations from complex environments and moving objects. Extensive experiments on three datasets demonstrated the effectiveness of our method, showcasing competitive performance.

\section*{Acknowledgements}
This work was supported in part by the National Key Research and Development Program of China under Grant 2021YFE0110500; in part by the National Natural Science Foundation of China under Grant 62062021; in part by the Beijing Natural Science Foundation (L231012). Yigang Cen is corresponding author.
{\small
\bibliographystyle{ieee_fullname}
\bibliography{PaperForReview}

\begin{thebibliography}{10}\itemsep=-1pt

\bibitem{Cai_Zhang_Liu_Gao_Hao_2022}
Ruichu Cai, Hao Zhang, Wen Liu, Shenghua Gao, and Zhifeng Hao.
\newblock Appearance-motion memory consistency network for video anomaly detection.
\newblock {\em Proceedings of the AAAI Conference on Artificial Intelligence}, page 938–946, Sep 2022.

\bibitem{Cao_Lu_Zhang_2022}
Congqi Cao, Yue Lu, and Yanning Zhang.
\newblock Context recovery and knowledge retrieval: A novel two-stream framework for video anomaly detection.
\newblock {\em arXiv preprint arXiv:2209.02899}, 2022.

\bibitem{chang2020clustering}
Yunpeng Chang, Zhigang Tu, Wei Xie, and Junsong Yuan.
\newblock Clustering driven deep autoencoder for video anomaly detection.
\newblock In {\em Computer Vision--ECCV 2020: 16th European Conference, Glasgow, UK, August 23--28, 2020, Proceedings, Part XV 16}, pages 329--345. Springer, 2020.

\bibitem{cong2011sparse}
Yang Cong, Junsong Yuan, and Ji Liu.
\newblock Sparse reconstruction cost for abnormal event detection.
\newblock In {\em CVPR 2011}, pages 3449--3456. IEEE, 2011.

\bibitem{dosovitskiy2020image}
Alexey Dosovitskiy, Lucas Beyer, Alexander Kolesnikov, Dirk Weissenborn, Xiaohua Zhai, Thomas Unterthiner, Mostafa Dehghani, Matthias Minderer, Georg Heigold, Sylvain Gelly, et~al.
\newblock An image is worth 16x16 words: Transformers for image recognition at scale.
\newblock {\em arXiv preprint arXiv:2010.11929}, 2020.

\bibitem{Feng_Song_Chen_Chen_Ni_Chen_2021}
Xinyang Feng, Dongjin Song, Yuncong Chen, Zhengzhang Chen, Jingchao Ni, and Haifeng Chen.
\newblock Convolutional transformer based dual discriminator generative adversarial networks for video anomaly detection.
\newblock In {\em Proceedings of the 29th ACM International Conference on Multimedia}, Oct 2021.

\bibitem{Georgescu_2021}
Mariana-Iuliana Georgescu, Antonio Barbalau, Radu~Tudor Ionescu, Fahad Shahbaz~Khan, Marius Popescu, and Mubarak Shah.
\newblock Anomaly detection in video via self-supervised and multi-task learning.
\newblock In {\em 2021 IEEE/CVF Conference on Computer Vision and Pattern Recognition (CVPR)}, Jun 2021.

\bibitem{Georgescu_Ionescu_Khan_Popescu_Shah_2023}
Mariana-Iuliana Georgescu, RaduTudor Ionescu, FahadShahbaz Khan, Marius Popescu, and Mubarak Shah.
\newblock A background-agnostic framework with adversarial training for abnormal event detection in video.
\newblock {\em IEEE Transactions on Pattern Analysis and Machine Intelligence,IEEE Transactions on Pattern Analysis and Machine Intelligence}, Jan 2023.

\bibitem{Gong_2019}
Dong Gong, Lingqiao Liu, Vuong Le, Budhaditya Saha, Moussa~Reda Mansour, Svetha Venkatesh, and Anton Van Den~Hengel.
\newblock Memorizing normality to detect anomaly: Memory-augmented deep autoencoder for unsupervised anomaly detection.
\newblock In {\em 2019 IEEE/CVF International Conference on Computer Vision (ICCV)}, Oct 2019.

\bibitem{Hasan_Choi_Neumann_Roy-Chowdhury_Davis_2016}
Mahmudul Hasan, Jonghyun Choi, Jan Neumann, Amit~K. Roy-Chowdhury, and Larry~S. Davis.
\newblock Learning temporal regularity in video sequences.
\newblock In {\em 2016 IEEE Conference on Computer Vision and Pattern Recognition (CVPR)}, Jun 2016.

\bibitem{he2016deep}
Kaiming He, Xiangyu Zhang, Shaoqing Ren, and Jian Sun.
\newblock Deep residual learning for image recognition.
\newblock In {\em Proceedings of the IEEE conference on computer vision and pattern recognition}, pages 770--778, 2016.

\bibitem{Huang_Zhao_Wang_Wu_2022}
Xiangyu Huang, Caidan Zhao, and Zhiqiang Wu.
\newblock A video anomaly detection framework based on appearance-motion semantics representation consistency.
\newblock In {\em ICASSP 2023-2023 IEEE International Conference on Acoustics, Speech and Signal Processing (ICASSP)}, pages 1--5. IEEE, 2023.

\bibitem{Ionescu_Smeureanu_Alexe_Popescu_2017}
Radu~Tudor Ionescu, Sorina Smeureanu, Bogdan Alexe, and Marius Popescu.
\newblock Unmasking the abnormal events in video.
\newblock In {\em 2017 IEEE International Conference on Computer Vision (ICCV)}, Oct 2017.

\bibitem{kim_cho_kweon_2019}
Dahun Kim, Donghyeon Cho, and In~So Kweon.
\newblock Self-supervised video representation learning with space-time cubic puzzles.
\newblock {\em Proceedings of the AAAI Conference on Artificial Intelligence}, page 8545–8552, Aug 2019.

\bibitem{kingma2014adam}
Diederik~P Kingma and Jimmy Ba.
\newblock Adam: A method for stochastic optimization.
\newblock {\em arXiv preprint arXiv:1412.6980}, 2014.

\bibitem{krizhevsky2012imagenet}
Alex Krizhevsky, Ilya Sutskever, and Geoffrey~E Hinton.
\newblock Imagenet classification with deep convolutional neural networks.
\newblock {\em Advances in neural information processing systems}, 25, 2012.

\bibitem{lin2014microsoft}
Tsung-Yi Lin, Michael Maire, Serge Belongie, James Hays, Pietro Perona, Deva Ramanan, Piotr Doll{\'a}r, and C~Lawrence Zitnick.
\newblock Microsoft coco: Common objects in context.
\newblock In {\em Computer Vision--ECCV 2014: 13th European Conference, Zurich, Switzerland, September 6-12, 2014, Proceedings, Part V 13}, pages 740--755. Springer, 2014.

\bibitem{liu2018future}
Wen Liu, Weixin Luo, Dongze Lian, and Shenghua Gao.
\newblock Future frame prediction for anomaly detection--a new baseline.
\newblock In {\em Proceedings of the IEEE conference on computer vision and pattern recognition}, pages 6536--6545, 2018.

\bibitem{liu2018ano_pred}
W. Liu, D.~Lian W.~Luo, and S. Gao.
\newblock Future frame prediction for anomaly detection -- a new baseline.
\newblock In {\em 2018 IEEE Conference on Computer Vision and Pattern Recognition (CVPR)}, 2018.

\bibitem{liu2021hybrid}
Zhian Liu, Yongwei Nie, Chengjiang Long, Qing Zhang, and Guiqing Li.
\newblock A hybrid video anomaly detection framework via memory-augmented flow reconstruction and flow-guided frame prediction.
\newblock In {\em Proceedings of the IEEE/CVF international conference on computer vision}, pages 13588--13597, 2021.

\bibitem{lu2013abnormal}
Cewu Lu, Jianping Shi, and Jiaya Jia.
\newblock Abnormal event detection at 150 fps in matlab.
\newblock In {\em Proceedings of the IEEE international conference on computer vision}, pages 2720--2727, 2013.

\bibitem{Lu_Shi_Jia_2013}
Cewu Lu, Jianping Shi, and Jiaya Jia.
\newblock Abnormal event detection at 150 fps in matlab.
\newblock In {\em 2013 IEEE International Conference on Computer Vision}, Dec 2013.

\bibitem{Luo_Liu_Gao_2017}
Weixin Luo, Wen Liu, and Shenghua Gao.
\newblock Remembering history with convolutional lstm for anomaly detection.
\newblock In {\em 2017 IEEE International Conference on Multimedia and Expo (ICME)}, Jul 2017.

\bibitem{Luo_Liu_Gao_2017_a}
Weixin Luo, Wen Liu, and Shenghua Gao.
\newblock A revisit of sparse coding based anomaly detection in stacked rnn framework.
\newblock In {\em 2017 IEEE International Conference on Computer Vision (ICCV)}, Oct 2017.

\bibitem{Lv_Chen_Cui_Xu_Li_Yang_2021}
Hui Lv, Chen Chen, Zhen Cui, Chunyan Xu, Yong Li, and Jian Yang.
\newblock Learning normal dynamics in videos with meta prototype network.
\newblock In {\em 2021 IEEE/CVF Conference on Computer Vision and Pattern Recognition (CVPR)}, Jun 2021.

\bibitem{5539872}
Vijay Mahadevan, Weixin Li, Viral Bhalodia, and Nuno Vasconcelos.
\newblock Anomaly detection in crowded scenes.
\newblock In {\em 2010 IEEE Computer Society Conference on Computer Vision and Pattern Recognition}, pages 1975--1981, 2010.

\bibitem{Morais_Le_Tran_Saha_Mansour_Venkatesh_2019}
Romero Morais, Vuong Le, Truyen Tran, Budhaditya Saha, Moussa Mansour, and Svetha Venkatesh.
\newblock Learning regularity in skeleton trajectories for anomaly detection in videos.
\newblock In {\em 2019 IEEE/CVF Conference on Computer Vision and Pattern Recognition (CVPR)}, Jun 2019.

\bibitem{morais2019learning}
Romero Morais, Vuong Le, Truyen Tran, Budhaditya Saha, Moussa Mansour, and Svetha Venkatesh.
\newblock Learning regularity in skeleton trajectories for anomaly detection in videos.
\newblock In {\em Proceedings of the IEEE/CVF conference on computer vision and pattern recognition}, pages 11996--12004, 2019.

\bibitem{Nguyen_Meunier_2019}
Trong~Nguyen Nguyen and Jean Meunier.
\newblock Anomaly detection in video sequence with appearance-motion correspondence.
\newblock In {\em 2019 IEEE/CVF International Conference on Computer Vision (ICCV)}, Oct 2019.

\bibitem{park2020learning}
Hyunjong Park, Jongyoun Noh, and Bumsub Ham.
\newblock Learning memory-guided normality for anomaly detection.
\newblock In {\em Proceedings of the IEEE/CVF conference on computer vision and pattern recognition}, pages 14372--14381, 2020.

\bibitem{Park_Noh_Ham_2020}
Hyunjong Park, Jongyoun Noh, and Bumsub Ham.
\newblock Learning memory-guided normality for anomaly detection.
\newblock In {\em 2020 IEEE/CVF Conference on Computer Vision and Pattern Recognition (CVPR)}, Jun 2020.

\bibitem{redmon2018yolov3}
Joseph Redmon and Ali Farhadi.
\newblock Yolov3: An incremental improvement.
\newblock {\em arXiv preprint arXiv:1804.02767}, 2018.

\bibitem{reiss2022attribute}
Tal Reiss and Yedid Hoshen.
\newblock Attribute-based representations for accurate and interpretable video anomaly detection.
\newblock {\em arXiv preprint arXiv:2212.00789}, 2022.

\bibitem{simonyan2014very}
Karen Simonyan and Andrew Zisserman.
\newblock Very deep convolutional networks for large-scale image recognition.
\newblock {\em arXiv preprint arXiv:1409.1556}, 2014.

\bibitem{Sun_Jia_Hu_Wu_2020}
Che Sun, Yunde Jia, Yao Hu, and Yuwei Wu.
\newblock Scene-aware context reasoning for unsupervised abnormal event detection in videos.
\newblock In {\em Proceedings of the 28th ACM International Conference on Multimedia}, Oct 2020.

\bibitem{szegedy2015going}
Christian Szegedy, Wei Liu, Yangqing Jia, Pierre Sermanet, Scott Reed, Dragomir Anguelov, Dumitru Erhan, Vincent Vanhoucke, and Andrew Rabinovich.
\newblock Going deeper with convolutions.
\newblock In {\em Proceedings of the IEEE conference on computer vision and pattern recognition}, pages 1--9, 2015.

\bibitem{tang2020integrating}
Yao Tang, Lin Zhao, Shanshan Zhang, Chen Gong, Guangyu Li, and Jian Yang.
\newblock Integrating prediction and reconstruction for anomaly detection.
\newblock {\em Pattern Recognition Letters}, 129:123--130, 2020.

\bibitem{wang2022video}
Guodong Wang, Yunhong Wang, Jie Qin, Dongming Zhang, Xiuguo Bao, and Di Huang.
\newblock Video anomaly detection by solving decoupled spatio-temporal jigsaw puzzles.
\newblock In {\em European Conference on Computer Vision}, pages 494--511. Springer, 2022.

\bibitem{Wang_Che_Jiang_Xiao_Yang_Tang_Ye_Wang_Qi_2020}
Xuanzhao Wang, Zhengping Che, Bo Jiang, Ning Xiao, Ke Yang, Jian Tang, Jieping Ye, Jingyu Wang, and Qi Qi.
\newblock Robust unsupervised video anomaly detection by multi-path frame prediction.
\newblock {\em Cornell University - arXiv,Cornell University - arXiv}, Nov 2020.

\bibitem{wang2020cluster}
Ziming Wang, Yuexian Zou, and Zeming Zhang.
\newblock Cluster attention contrast for video anomaly detection.
\newblock In {\em Proceedings of the 28th ACM international conference on multimedia}, pages 2463--2471, 2020.

\bibitem{Wu_Liu_Shen_2019}
Peng Wu, Jing Liu, and Fang Shen.
\newblock A deep one-class neural network for anomalous event detection in complex scenes.
\newblock {\em IEEE Transactions on Neural Networks and Learning Systems}, page 1–14, Jan 2019.

\bibitem{Yang_Liu_Wu_Wu_Liu_2023}
Zhiwei Yang, Jing Liu, Zhaoyang Wu, Peng Wu, and Xiaotao Liu.
\newblock Video event restoration based on keyframes for video anomaly detection.
\newblock In {\em Proceedings of the IEEE/CVF Conference on Computer Vision and Pattern Recognition}, pages 14592--14601, 2023.

\bibitem{yang2022dynamic}
Zhiwei Yang, Peng Wu, Jing Liu, and Xiaotao Liu.
\newblock Dynamic local aggregation network with adaptive clusterer for anomaly detection.
\newblock In {\em European Conference on Computer Vision}, pages 404--421. Springer, 2022.

\bibitem{Ye_Peng_Gan_Wu_Qiao_2019}
Muchao Ye, Xiaojiang Peng, Weihao Gan, Wei Wu, and Yu Qiao.
\newblock Anopcn.
\newblock In {\em Proceedings of the 27th ACM International Conference on Multimedia}, Oct 2019.

\bibitem{ye2019anopcn}
Muchao Ye, Xiaojiang Peng, Weihao Gan, Wei Wu, and Yu Qiao.
\newblock Anopcn: Video anomaly detection via deep predictive coding network.
\newblock In {\em Proceedings of the 27th ACM International Conference on Multimedia}, pages 1805--1813, 2019.

\bibitem{Yu_Wang_Cai_Zhu_Xu_Yin_Kloft_2020}
Guang Yu, Siqi Wang, Zhiping Cai, En Zhu, Chuanfu Xu, Jianping Yin, and Marius Kloft.
\newblock Cloze test helps: Effective video anomaly detection via learning to complete video events.
\newblock In {\em Proceedings of the 28th ACM International Conference on Multimedia}, Oct 2020.

\bibitem{zhao2017spatio}
Yiru Zhao, Bing Deng, Chen Shen, Yao Liu, Hongtao Lu, and Xian-Sheng Hua.
\newblock Spatio-temporal autoencoder for video anomaly detection.
\newblock In {\em Proceedings of the 25th ACM international conference on Multimedia}, pages 1933--1941, 2017.

\bibitem{Zhong_Chen_Jiang_Ren_2022}
Yuanhong Zhong, Xia Chen, Jinyang Jiang, and Fan Ren.
\newblock A cascade reconstruction model with generalization ability evaluation for anomaly detection in videos.
\newblock {\em Pattern Recognition}, page 108336, Feb 2022.

\bibitem{Zhou_Du_Zhu_Peng_Liu_Goh_2019}
Joey~Tianyi Zhou, Jiawei Du, Hongyuan Zhu, Xi Peng, Yong Liu, and Rick Siow~Mong Goh.
\newblock Anomalynet: An anomaly detection network for video surveillance.
\newblock {\em IEEE Transactions on Information Forensics and Security}, page 2537–2550, Oct 2019.

\end{thebibliography}
}
\end{document}